\documentclass[letterpaper]{article} 
\usepackage{aaai25}  
\usepackage{times}  
\usepackage{helvet}  
\usepackage{courier}  
\usepackage[hyphens]{url}  
\usepackage{graphicx} 
\usepackage{natbib}  
\usepackage{caption} 
\usepackage{algorithm}
\usepackage{algorithmic}
\usepackage{multirow}
\usepackage{amsmath,amsthm,amssymb,amsfonts}
\usepackage{graphicx}
\usepackage{subcaption}
\usepackage{booktabs}

\usepackage{newfloat}
\usepackage{listings}
\DeclareCaptionStyle{ruled}{labelfont=normalfont,labelsep=colon,strut=off} 
\floatstyle{ruled}
\newfloat{listing}{tb}{lst}{}
\floatname{listing}{Listing}
\title{Digging into Intrinsic Contextual Information for \\ High-fidelity 3D Point Cloud Completion}
\author {
    Jisheng Chu\textsuperscript{\rm 1},
    Wenrui Li\textsuperscript{\rm 1},
    Xingtao Wang\textsuperscript{\rm 1, \rm 3}\thanks{Corresponding author.},
    Kanglin Ning\textsuperscript{\rm 1, \rm 3},
    Yidan Lu\textsuperscript{\rm 1} and Xiaopeng Fan\textsuperscript{\rm 1, \rm 2, \rm 3}
}
\affiliations {
    \textsuperscript{\rm 1}Harbin Institute of Technology \textsuperscript{\rm 2}Pengcheng Laboratory \textsuperscript{\rm 3}Harbin Institute of Technology Suzhou Research Institute\\
    23B936009@stu.hit.edu.cn; liwr618@163.com; xtwang@hit.edu.cn;\\23B936010@stu.hit.edu.cn; 24S103311@stu.hit.edu.cn; fxp@hit.edu.cn 
}

\usepackage{bibentry}
\begin{document}

\maketitle

\begin{abstract}
The common occurrence of occlusion-induced incompleteness in point clouds has made point cloud completion (PCC) a highly-concerned task in the field of geometric processing. 
Existing PCC methods typically produce complete point clouds from partial point clouds in a coarse-to-fine paradigm, with the coarse stage generating entire shapes and the fine stage improving texture details. Though diffusion models have demonstrated effectiveness in the coarse stage, the fine stage still faces challenges in producing high-fidelity results due to the ill-posed nature of PCC. The intrinsic contextual information for texture details in partial point clouds is the key to solving the challenge.
In this paper, we propose a high-fidelity PCC method that digs into both short and long-range contextual information from the partial point cloud in the fine stage.
Specifically, after generating the coarse point cloud via a diffusion-based coarse generator, a mixed sampling module introduces short-range contextual information from partial point clouds into the fine stage. A surface freezing modules safeguards points from noise-free partial point clouds against disruption.
As for the long-range contextual information, we design a similarity modeling module to derive similarity with rigid transformation invariance between points, conducting effective matching of geometric manifold features globally.
In this way, the high-quality components present in the partial point cloud serve as valuable references for refining the coarse point cloud with high fidelity.
Extensive experiments have demonstrated the superiority of the proposed method over SOTA competitors. Our code is available at https://github.com/JS-CHU/ContextualCompletion.
\end{abstract}

%

\section{Introduction}

Point cloud, a widely used representation for object geometry in 3D space, offers a simple and flexible data structure. However, raw point clouds obtained through devices like laser scanners, often exhibit missing regions due to factors such as occlusion, surface reflectivity, and scanning range limitations. The incompleteness of  point clouds adversely affects 3D model quality and effectiveness of higher-level tasks like classification, segmentation, and object detection. Consequently, point cloud completion (PCC), the task filling in missing regions given partial point clouds, is of paramount importance.

\begin{figure}[H]
\centering
\includegraphics[width=0.999\columnwidth]{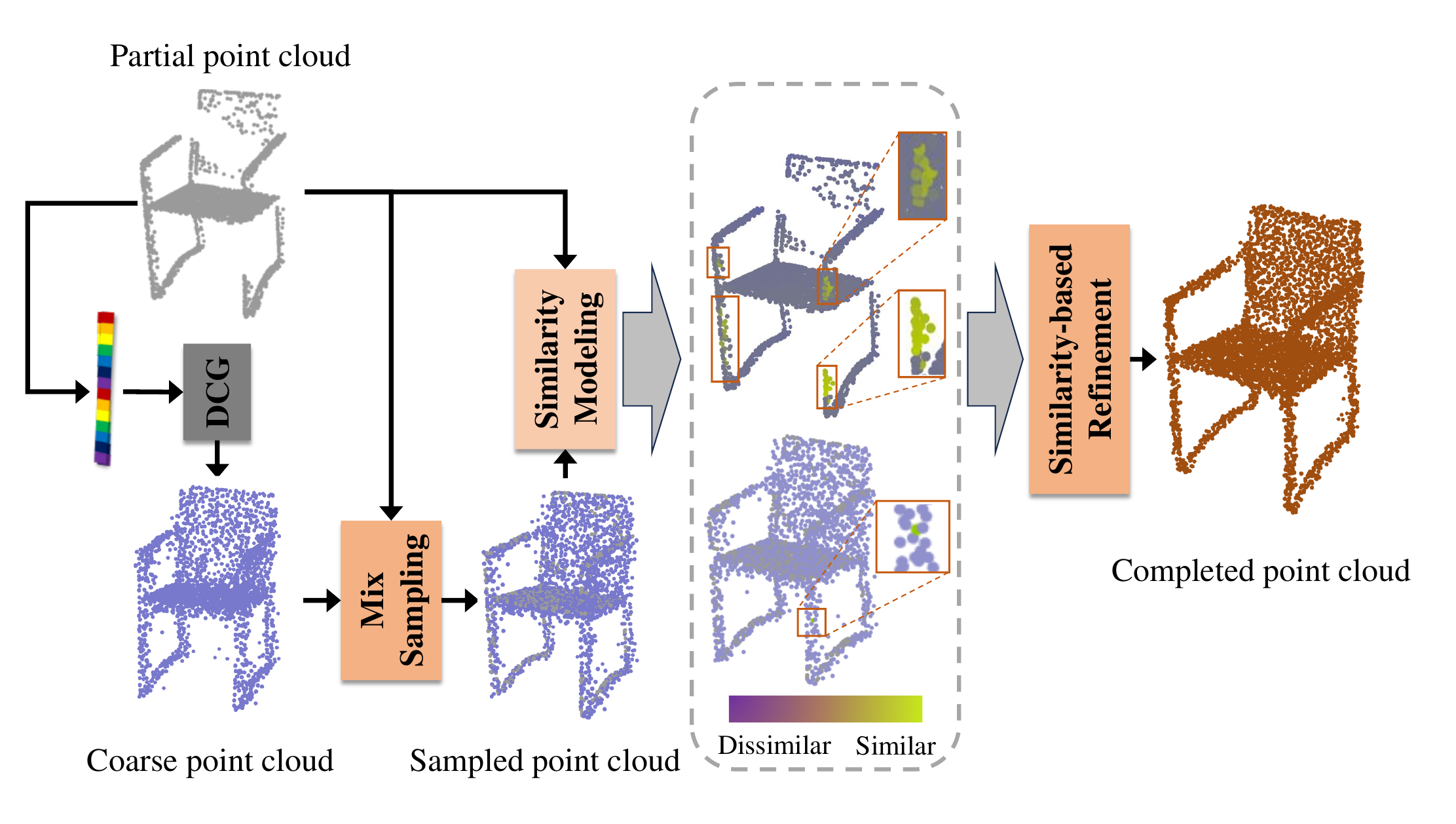} 
\caption{The workflow of the proposed method. In the middle of the figure, we visualize the matching of non-local regions based on the similarity of geometric structures. A higher degree of similarity is represented by a more intense yellow color. The green point in the sampled point cloud is refined refering to the heatmap in the partial point cloud.}
\label{figure:workflow}
\end{figure}

With advancements in deep learning, notable progress has been achieved in the field of PCC. Existing PCC networks \cite{Yuan-PCN,Zhang-vipc,Ma-USSPA} generally produce complete point clouds from partial ones following a coarse-to-fine paradigm. The coarse stage produces a preliminary shape based on learned prior knowledge from training samples, and the fine stage enhances geometric texture details. Recently, diffusion probabilistic models \cite{Luo-DPM} have been extended to address the coarse stage, yielding impressive performance \cite{Lyu-Cdiffusion}. However, the fine stage still faces challenges as recovering the miss regions is actually an ill-posed problem. Fortunately, real-world objects often exhibit symmetries and regularities, making the intrinsic contextual information of texture details within partial point clouds invaluable for refining the coarse point cloud with high fidelity.

In this paper, we propose a high-fidelity PCC method adhering to a two-stage completion strategy: coarse point cloud generation followed by a refinement network. The proposed method digs into both short-range and long-range contextual information from the partial point cloud in a proposed Context-aware Refiner (CRef). As shown in Figure \ref{figure:workflow}, taking a global feature of the partial point cloud as the condition, a diffusion model is trained to generate the preliminary coarse point cloud. Subsequently, a mixed sampling module is introduced to seamlessly merge the coarse point cloud with the partial point cloud. This integration serves to incorporate short-range contextual details from the partial point clouds into the fine stage. Within this module, a surface freezing mechanism is implemented to safeguard the points originating from noise-free partial point clouds against disturbances during the fine stage.
As for the long-range contextual information, a similarity modeling module is designed, incorporating rotation matrix and symmetric plane learning, to facilitate robust global matching of rigid transformation-invariant geometric manifold features. In this way, the high-quality components present in the partial point cloud can serve as valuable references for refining the coarse point cloud with high fidelity. Extensive experiments covering synthetic and real-scanned data have demonstrated the superiority of our method over its competitors.

The contributions can be summarized as follows:
\begin{itemize}
    \item We propose a high-fidelity method for 3D point cloud completion, digging into both short-range and long-range contextual information from the partial point cloud for high-fidelity refinement.
    \item We design a mixed sampling module and surface freezing mechanism, incorporating short-range contextual details from the partial point clouds into the fine stage.
    \item We design a learnable rigid transformation and a similarity modeling module to extract long-range contextual information, which conducts effective matching of rigid transformation-invariant geometric manifold features globally.
\end{itemize}

\section{Related Works}
\begin{figure*}[t]
\centering
\includegraphics[width=0.99\textwidth]{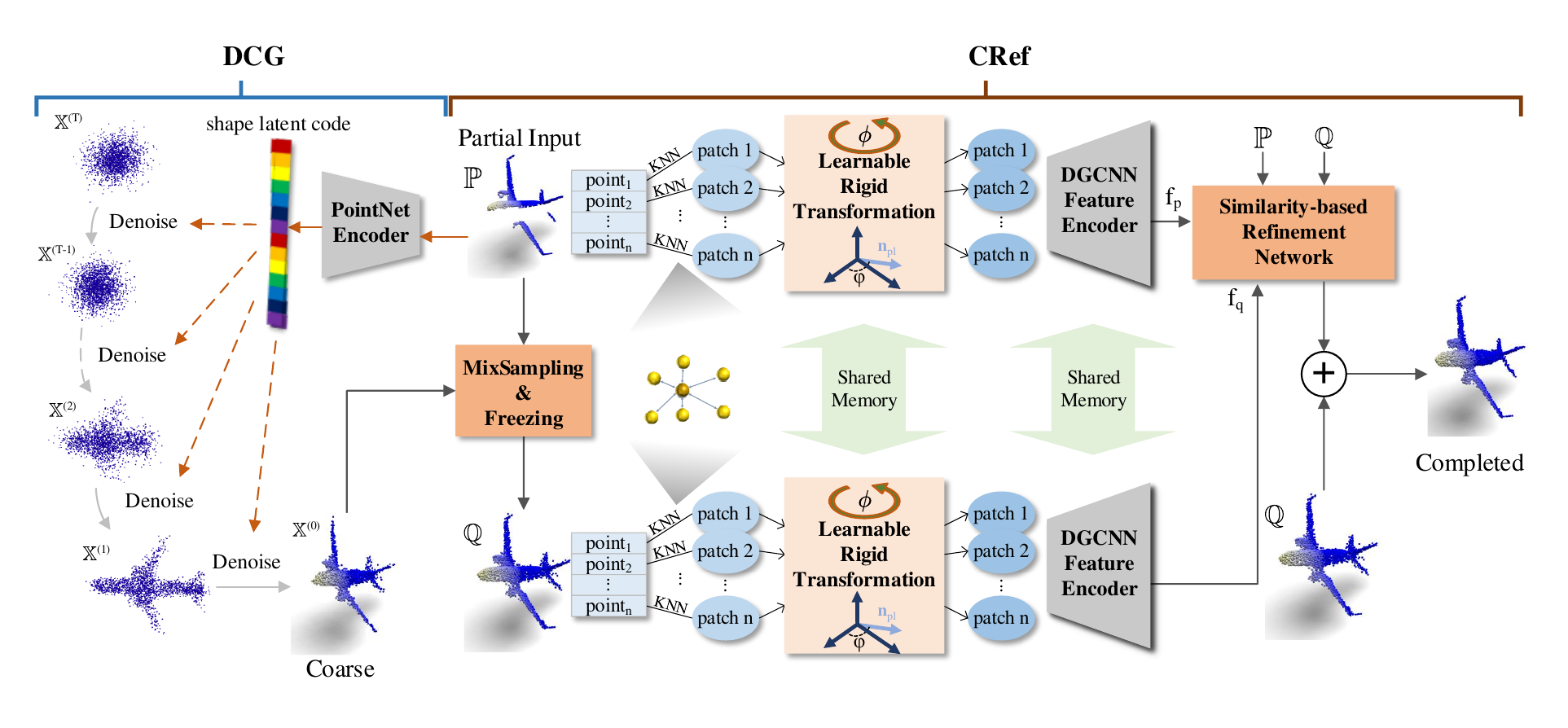} 
\caption{The overall architecture consists of a Diffusion-based Coarse Generator (DCG) and a Context-aware Refiner (CRef). In DCG, a PointNet Encoder extracts a  global shape latent code from the partial input as a condition. The coarse point cloud is generated through denoising process. The coarse point cloud is then refined in CRef according to both short and long-range contextual information, deriving a point cloud with entire shape and high-fidelity textural details.}
\label{figure:structure}
\end{figure*}
Benefiting from advancements in deep learning, The domains of image processing \cite{li2024multiscale,chen2022geometry,li2024spiking,chen2025divide}, video analysis \cite{li2024ustc,tang2024offline}, point cloud processing \cite{PointFilterNet,FCNet}, and multimodal technologies \cite{li2024multilayer,li2022neuron} are experiencing unprecedented advancements at an accelerated pace. In the field of point cloud completion (PCC), substantial progress has been achieved. Several methods \cite{Dai-2017,Xie-GRNet} map the points to voxels and employ 3D convolution networks on these voxels. However, completion on voxel-level involves expensive computational cost, and the voxelization of points results in the loss of surface texture details.

Since the success of point cloud analysis methods \cite{Qi-PointNet, Qi-PointNet++,Li-PointCNN,Wu-PointConv,Wang-DGCNN,Zhao-PointTransformer}, PCN \cite{Yuan-PCN} is the pioneering point-level method to employ an encoder-decoder architecture for PCC. It utilizes a simple network to generate a coarse point cloud and conduct completion following the FoldingNet \cite{Yang-FoldingNet}. TopNet \cite{Tchapmi-TopNet} designes a decoder with a tree structure implying local information to generate structured point clouds.
PF-Net \cite{Huang-PF-Net} utilizes GAN architecture and proposes a multi-resolution encoder and a pyramid decoder to predict points with hierarchical global details. ViPC \cite{Zhang-vipc} and CSDN \cite{Zhu-CSDN} incorporate global features from both point cloud and image modalities and obtain global constraints from the two modalities. USSPA \cite{Ma-USSPA}, based on a GAN network framework, introduces a symmetrical learning module to learn and leverage symmetrical information from partial point clouds.
The rise in popularity of transformers has catalyzed PointTr \cite{Yu-PointTr}, which redefines PCC as a set-to-set translation problem. It introduces a novel encoder-decoder architecture based on transformers and integrates a geometry-aware module to explicitly model local geometric relationships. XMF-net \cite{Aiello-XMF-net} stacks multiple self attention modules and cross modules, integrating information from both image and point cloud modality. VoxFormer \cite{Li-VoxFormer} utilizes the transformer architecture for voxels, first estimating depth from images, and then integrating image features to reconstruct point clouds. To integrate cross-resolution point cloud features, CRA-PCN \cite{Rong-CRA_PCN} efficiently leverages local attention mechanisms for high-resolution aggregation and switches inputs to perform intra-layer or inter-layer cross-resolution aggregation. Inspired by PointTr, ProxyFormer \cite{Li-ProxyFormer} introduces existing proxies and missing proxies to represent features of existing and missing parts. It aims to generate only the missing parts to complete the restoration, supervised by real missing parts. 

Recently the remarkable success of diffusion models in image generation has brought to point cloud generation \cite{Luo-DPM,zeng-LION,Nakayama-DiffFacto} and achieved significant advancements. Diffusion-based methods have begun to emerge in the field of PCC. PVD \cite{PVD} encodes the partial input as condition, conducting diffusion model on voxel-level. PDR \cite{Lyu-Cdiffusion} extracts multi-level features from the partial input as conditions. With a novel dual-path architecture for diffusion and refinement networks, PDR achieves excellent results in both coarse point cloud generation and completion. The multimodal diffusion-based completion methods \cite{Cheng-SDFusion,Kasten-TtIDiffusion} utilize image and text respectively as additional modalities to control the reverse diffusion process. DiffComplete \cite{DiffComplete} proposes a hierarchical feature aggregation strategy to control the outputs for a single condition and introduces an occupancy-aware fusion strategy to incorporate more shape details for multiple conditions. 

Existing coarse-to-fine methods typically focus on both local and global features. However, the local features in coarse point clouds are often imprecise. Additionally, previous methods do not account for the rigid transformation-invariant feature similarities between local regions, leading to insufficient utilization of intrinsic contextual information in the partial point cloud. Our method addresses these problems to effectively obtain high-fidelity completion results.

\section{Methodology}
The proposed PCC method follows the coarse-to-fine paradigm, which takes a diffusion-based coarse generator (DCG) for coarse generation and digs into intrinsic contextual information from the partial point cloud in the Context-aware Refiner (CRef).

The pipeline of the proposed method is illustrated in Figure \ref{figure:structure}. Given a partial point cloud $\mathbb P$, DCG generates a coarse point cloud $\mathbb P_{coarse}$ that exhibits the entire shape with poor textual details.
$\mathbb P_{coarse}$ is then refined in CRef according to both short and long-range contextual information, deriving a point cloud with entire shape and high fidelity.

In the following, we will provide a brief introduction to DCG, followed by detailed descriptions of CRef.

\subsection{Diffusion-based Coarse Generator (DCG)}

We provide a brief overview of DCG, which is heavily inspired by the model proposed by Luo \cite{Luo-DPM}.

As shown in Figure \ref{figure:structure}, DCG employs a PointNet as an encoder to extract a global shape latent code from the partial input. Subsequently, utilizing the global code as a condition, the reverse diffusion sampling process is controlled to generate a complete coarse point cloud from Gaussian noise. 

The diffusing process can be formalized as:
\begin{equation}
q( \mathbb{X}_{1:T}| \mathbb{X}_{0})=\prod_{t=1}^{T} q( \mathbb{X}_t| \mathbb{X}_{t-1}),
\end{equation}
\begin{equation}
q( \mathbb{X}_t| \mathbb{X}_{t-1})= N( \mathbb{X}_t|\sqrt{1-\beta_t} \mathbb{X}_{t-1},\beta_t\mathbf I).
\end{equation}

The reverse sampling process can be formalized as:
\begin{equation}
p_{\theta}( \mathbb{X}_{0:T}|z)=p( \mathbb{X}_{T}) \prod_{t=1}^{T} p_{\theta}( \mathbb{X}_{t-1}| \mathbb{X}_t,z),
\end{equation}
\begin{equation}
p_{\theta}( \mathbb{X}_{t-1}| \mathbb{X}_t,z)= N( \mathbb{X}_{t-1}|\mu_{\theta}( \mathbb{X}_t, t,z),\beta_t\mathbf I).
\end{equation}

In this setup, where $z$ represents the shape latent code extracted from the partial point cloud by the feature encoder, $\mathbb{X}_0$ denotes the ground truth point cloud, and $\mathbb{X}_T$ represents the Gaussian noise formed after $T$ steps of diffusion. The training objective aims to maximize the lower bound of the log-likelihood: ${E}_q[\log p_\theta( \mathbb{X}_0| \mathbb{X}_T,z)]$, which is operationalized by minimizing the Mean Squared Error (MSE) loss between $\mu_{\theta}$ and $\mu$ of the standard normal distribution.

\subsection{Context-aware Refiner (CRef)}
In this section, we present a comprehensive description of CRef, comprising three primary modules. The Short-range Contextual Information Extraction derives precise local manifold structures, while the Long-range Contextual Information Extraction captures rigid transformation-invariant features of local patches. Based on the obtained intrinsic contextual features, a refinement network is proposed to refine the coarse point cloud by similarity in both euclidean and feature space between non-local regions.
\subsubsection{Short-range Contextual Information Extraction.}
Precise short-range contextual information exists within the local space of the partial point cloud. A \textbf{mixed sampling module} integrates high-quality surface information into the fine stage by combining the coarse point cloud with the partial one. Meanwhile, a \textbf{surface freezing module} is devised to preserve the precise distribution of surface in the original partial point cloud.

Technically, after generating $\mathbb{P}_{coarse}$, we concatenate $\mathbb{P}$ and $\mathbb{P}_{coarse}$ to form a point cloud denoted as $\mathbb{P}_{concat}$ and then perform farthest point sampling \cite{Qi-PointNet++} on $\mathbb{P}_{concat}$. Subsequently, the surface freezing module identifies points that are from $\mathbb{P}$ and immobilizes them, marking those points with deviations as adjustable. This process can be conducted as follows:
\begin{equation}
\mathbb Q=SurfaceFreezing(FPS(\mathbb P\cup \mathbb P_{coarse},n)), 
\end{equation}
where $n$ stands for the number of points in $\mathbb Q$. $FPS(\cdot)$ represents farthest point sampling.

\begin{figure}[t]
\centering
\includegraphics[width=0.96\columnwidth]{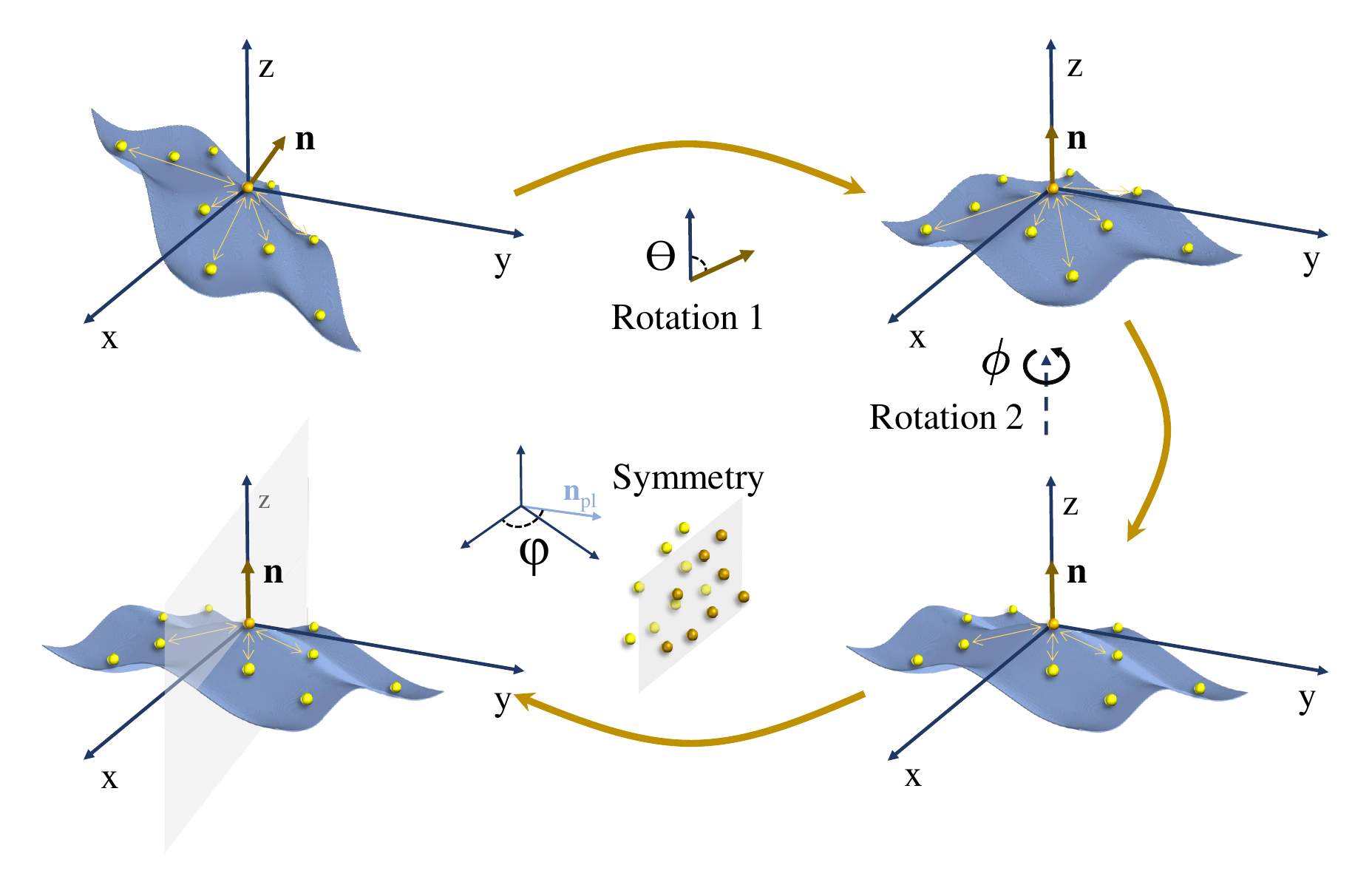} 
\caption{A patch is firstly rotated to a certain direction by $\theta$ and secondly rotated by the learned angle $\phi$. Finally, the patch performs symmetry with respect to the symmetry plane defined by the learned angle $\psi$.}
\label{figure:rigid}
\end{figure}

\subsubsection{Long-range Contextual Information Extraction.}

Long-range contextual information can be obtained by measuring the similarity between geometric features that remain invariant under rigid transformations. For the patches in $\mathbb{P}_{coarse}$, similar geometric manifold structures hide in $\mathbb{P}$, analogous to completing the right engine of an airplane by referencing the left engine. It's evident that these manifold structures can be corresponded through rigid transformations such as rotation, symmetry, etc. We propose a learnable rigid transformation which contains the rotation matrix learning and the symmetric plane learning.

For each patch $\mathbf{P}_i$ in $\mathbb{P}$ and $\mathbf{Q}_j$ in $\mathbb{Q}$, the coordinates of each point in the patch are subtracted by the centroid of the patch, and the number of points within a patch is denoted as ${k}$. We utilize the K-nearest neighbor ($KNN$) algorithm to extract patches:
\begin{equation}
\mathbf{P}_i \mathop{\leftarrow}\limits ^{i}  {KNN}(\mathbb P,  k),\ \ \mathbf{Q}_j \mathop{\leftarrow}\limits ^{j}  {KNN}( \mathbb Q,  k),
\end{equation}
where $\mathop{\leftarrow}\limits ^{i}$ indicates retrieving the $i^{th}$ patch from output of ${KNN}$. The normal vector of patch $ \mathbf{P}_i$, $ \mathbf{Q}_j$ are denoted as $\mathbf{n}_{p_i}$, and $\mathbf{n}_{q_j}$. To obtain features invariant to rigid transformations, we first apply rotation and symmetry transformations. 

As shown in Figure \ref{figure:rigid}, we firstly rotate $ \mathbf{P}_i$ and $\mathbf{Q}_j$ to the direction where their point normals are both $\mathbf{ e}_z=(0,0,1)^T$. The rotation matrix $ \mathbf{R}_{p_i}^{(1)}$ and  $ \mathbf{R}_{q_j}^{(1)}$ can be calculated using Rodrigues' rotation formula:
\begin{equation}
\mathbf{ k} = \frac{\mathbf{ n} \times \mathbf{ e}_z}{\|\mathbf{ n} \times \mathbf{ e}_z\|},
\label{formula:axis_angle_k}
\end{equation}
\begin{equation}
\theta=\arccos (\mathbf{ n}\cdot\mathbf{ e}_z),
\label{formula:axis_angle_theta}
\end{equation}
\begin{equation}
 \mathbf{R}^{(1)} = \cos(\theta) {\mathbf I} + (1 - \cos(\theta)) \mathbf{ k}\mathbf{ k}^T + \sin(\theta) [\mathbf{ k}]_{\times},
\label{formula:R_matrix1}
\end{equation}
where $\mathbf{ k}$ denotes the rotation axis from vector $\mathbf{ n}$ to vector $\mathbf{ e}_z$, $\mathbf{I}$ denotes an identity matrix, and $[\mathbf{ k}]_{\times}$ denotes skew-symmetric matrix of axis $\mathbf{ k}$. Then the rotated patches can be calculated as follows:
\begin{equation}
 \mathbf{\tilde {P}}_i \gets ( \mathbf{R}_{p_i}^{(1)} \cdot  \mathbf{P}_i^T)^T,\ \ \mathbf{\tilde {Q}}_j \gets ( \mathbf{R}_{q_j}^{(1)} \cdot  \mathbf{Q}_j^T)^T.
\end{equation}

$ \mathbf{\tilde {P}}_i$ and $ \mathbf{\tilde {Q}}_j$ are aligned in the same direction after the above transformation. However, in 3D space, the rotation angle around the axis $\mathbf{ e}_z$ is another degree of freedom affecting the alignment. Consequently, a MLP-form rotation learning network is deployed to learn the rotation angle $\phi$ around the axis $\mathbf{e}_z$. The rotation matrix can be calculated as:
\begin{equation}
 \mathbf{R}^{(2)} = \cos(\phi) {\mathbf I} + (1 - \cos(\phi)) \mathbf{e}_z\mathbf{e}_z^T + \sin(\phi) [\mathbf{e}_z]_{\times}.
\label{formula:R_matrix2}
\end{equation}

Getting another pair of rotation matrix $ \mathbf{R}_{p_i}^{(2)}$ and $ \mathbf{R}_{q_j}^{(2)}$, the final rotated patches can be calculated as follows:
\begin{equation}
\mathbf{\hat P}_i \gets ( \mathbf{R}_{p_i}^{(2)} \cdot  \mathbf{\tilde P}_i^T)^T,\ \ \mathbf{\hat Q}_j \gets ( \mathbf{R}_{q_j}^{(2)} \cdot  \mathbf{\tilde Q}_j^T)^T.
\end{equation}

The rotation transformation step aligns the patch $\mathbf{\hat P}_i$ and $\mathbf{\hat Q}_j$ to $\mathbf{{e}}_z$ with reference to the normal of the center point. The symmetric plane $\mathcal{M}$ passes through the center point, which indicates that $\mathbf{{e}}_z$ must lie on $\mathcal{M}$. In other words, the normal vector of $\mathcal{M}$, denoted as $\mathbf{{n}}_{pl}$, must be perpendicular to $\mathbf{{e}}_z$. Therefore, only one degree of freedom needs to be determined: the angle $\psi$ between $\mathbf{{n}}_{pl}$ and the x-axis. $\mathbf{{n}}_{pl}$ is computed as $\mathbf{{n}}_{pl} = (\cos(\psi),\sin(\psi),0)^T$, while points $\mathbf{\hat p}_i$ and $\mathbf{\hat q}_j$ after symmetry can be computed as:
\begin{equation}
\mathbf{p^{\prime}}_i \gets  \mathbf{\hat p}_i - 2\dfrac{\mathbf{ n}_{pl}^T \cdot  \mathbf{\hat p}_i}{\|\mathbf{ n}_{pl}^T\|^2} \mathbf{ n}_{pl}^T,\ \ \mathbf{q^{\prime}}_j \gets  \mathbf{\hat q}_j - 2\dfrac{\mathbf{ n}_{pl}^T \cdot  \mathbf{\hat q}_j}{\|\mathbf{ n}_{pl}^T\|^2} \mathbf{ n}_{pl}^T.
\end{equation}

For this purpose, we design a network structure to learn the rotation angle around the rotation axis in the rotation transformation part. Furthermore, parameters are shared in the first few layers to extract similar geometric features.

After rigid transformations involving rotation and symmetry, we achieve alignment of the patch level in the coordinate space. Subsequently, a simplified Dynamic Graph CNN (DGCNN) \cite{Wang-DGCNN} network is applied to extract features of each point $\mathbf{p^{\prime}}_i$ and $\mathbf{q^{\prime}_j}$ from the transformed patches. The original DGCNN network dynamically updates the graph model at each layer. Thanks to the rigid transformation module proposed in our work, we can simplify the DGCNN by updating the graph model only in the first and last layers to reduce computational costs. 
\begin{figure*}[t]
\centering
\includegraphics[width=0.99\textwidth]{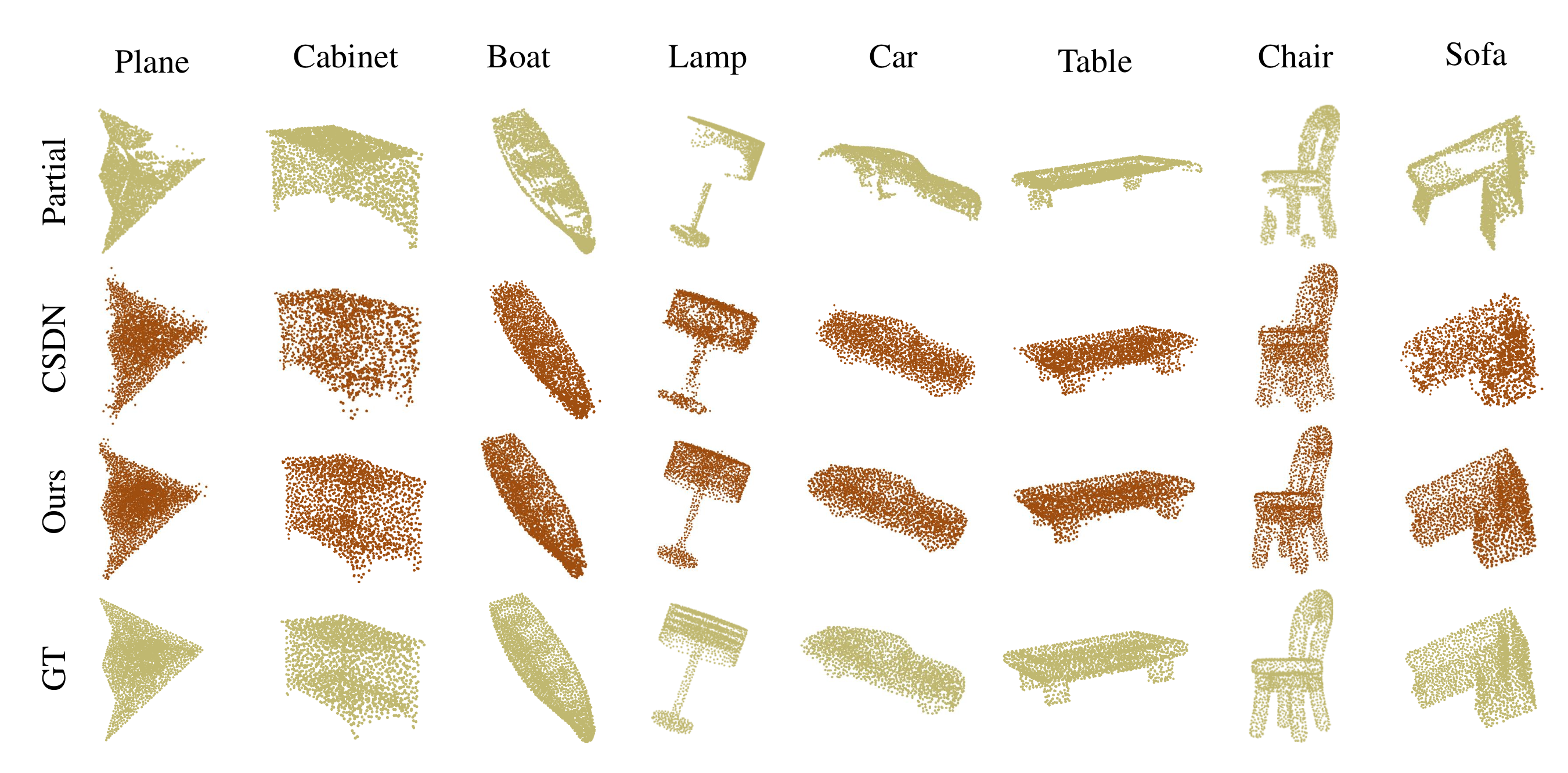} 
\caption{Qualitative comparison on ShapeNet-ViPC. The resolution for all point clouds are 2,048.}
\label{figure:visual}
\end{figure*}
\subsubsection{Non-local Similarity-based Refinement Network.}We measure both the Euclidean  similarity and feature space similarity between patches from $\mathbb{P}$ and $\mathbb{Q}$, utilizing the rich short-range and long-range contextual information to guide the displacement of points in $\mathbb{Q}$.

In coordinate space, points that are close in distance typically share similar geometric manifold structures, and we quantify this similarity using Euclidean distance:
\begin{equation}
 \mathbf{w}_1^j = (\|\mathbf{q}_j -  \mathbf{p}_1\|_2^2,\| \mathbf{q}_j -  \mathbf{p}_2\|_2^2,\dots,\| \mathbf{q}_j -  \mathbf{p}_{m}\|_2^2),
\end{equation}
where $\mathbf{p}_1$ and $\mathbf{q}_j$ are points from $ \mathbf{P}_i$ and $\mathbf{Q}_j$ before rigid transformation. Patches with analogous geometric manifold structures are mapped to nearby positions in high-dimensional feature space. We measure this similarity by Calculate the cosine distance between the patch feature $\mathbf{f}_q^j$ and $\mathbf{f}_p^i$:
\begin{equation}
 \mathbf{w}_2^j = ({cos}( \mathbf{f}_q^j, \mathbf{f}_p^1),{cos}(\mathbf{f}_q^j, \mathbf{f}_p^2),\dots,{cos}(\mathbf{f}_q^j, \mathbf{f}_p^{m}).
\end{equation}

At last, we perform element-wise multiplication on $\mathbf{w}_1$ and $\mathbf{w}_2$ to measure the similarity:
\begin{equation}
 \mathbf{W} =  e^{-\mathbf{W}_1} + e^{\mathbf{W}_2},
\end{equation}
where the $ \mathbf{W}$ we retain is a matrix of size $n \times m$. Each row of $ \mathbf{W} $ represents the similarity between a point in $\mathbb{Q}$ and all points in $\mathbb{P}$. We select the top $ k $ similarities and aggregate the corresponding similar features in $\mathbf{F}_p$. Then, we perform average pooling and max pooling on these similar features, concatenate them with $\mathbf{F}_{q}$, and construct the fused features. It can be formalized as follows:
\begin{equation}
 \mathbf{F} = {Aggregate}({TopK}(\mathbf{W}), \mathbf{F}_p),
\end{equation}
\begin{equation}
\mathbf {F} \gets \mathbf{F}_q\cup{MaxPool}(\mathbf F)\cup{AvgPool}(\mathbf F).
\label{formula:featurefuse}
\end{equation}

For every unfrozen point $\mathbf q_j$ in $\mathbb{Q}$, the fused feature $\mathbf{f}_j$ can be obtained according to Eq. \ref{formula:featurefuse}. We conduct a MLP to learn the displacement $\mathbf{o}_j$ for $\mathbf{q}_j$. Then we rotate this displacement vector back as:
\begin{equation}
 {\mathbf{o}_j} \gets ({\mathrm{R}^{(2)}}^{-1}({\mathrm{R}^{(1)}}^{-1}\mathbf{o}_j^T))^T.
\end{equation}

Finally we add $\mathbf{o}_j$ to $\mathbf{q}_j$ to get the refined point:
\begin{equation}
 \mathbf{{q^*}}_j \gets \mathbf{q}_j + { \mathbf{o}_j}.
\end{equation}
  
\section{Experiments}
In this section, we first introduce experimental settings and five datasets in details, and then present extensive experiments including comparison study with existing methods and ablation study.
\begin{table*}[t]
\centering
\begin{tabular}{c|c|cccc|c|cccc}
\toprule[1.2pt]
\multirow{2}{*}{Methods} & \multicolumn{5}{c}{$L_2$ CD $\times 10^{-3}$} & \multicolumn{5}{c}{F-score@0.001} \\ 
 & Avg. & Airplane & Chair & Lamp & Watercraft & Avg. & Airplane & Chair & Lamp & Watercraft \\ \midrule[0.5pt]
PCN \shortcite{Yuan-PCN} & 5.619 & 4.246 & 7.441 & 6.331 & 3.510 & 0.407 & 0.578 & 0.323 & 0.456 & 0.577 \\
TopNet \shortcite{Tchapmi-TopNet} & 4.976 & 3.710 & 6.391 & 5.547 & 3.350 & 0.467 & 0.593 & 0.388 & 0.491 & 0.615 \\
GRNet \shortcite{Xie-GRNet} & 3.171 & 1.916 & 3.402 & 3.034 & 2.160 & 0.601 & 0.767 & 0.575 & 0.694 & 0.704 \\
PF-Net \shortcite{Huang-PF-Net} & 3.873 & 2.515 & 4.478 & 5.185 & 2.871 & 0.551 & 0.718 & 0.489 & 0.559 & 0.656 \\
PoinTr \shortcite{Yu-PointTr} & 2.851 & 1.686 & 3.111 & 2.928 & 1.737 & 0.683 & 0.842 & 0.662 & 0.742 & 0.780 \\
ViPC \shortcite{Zhang-vipc} & 3.308 & 1.760 & 2.476 & 2.867 & 2.197 & 0.591 & 0.803 & 0.529 & 0.706 & 0.730 \\
Seedformer \shortcite{Zhou-SeedFormer} & 2.902 & 1.716 & 3.151 & 3.226 & 1.679 & 0.688 & 0.835 & 0.668 & 0.777 & 0.786 \\
CSDN \shortcite{Zhu-CSDN} & 2.570 & 1.251 & 2.835 & 2.554 & 1.742 & 0.695 & 0.862 & 0.669 & 0.761 & 0.782 \\
PointAttN \shortcite{PointAttN} & 2.853 & 1.613 & 3.157 & 3.058 & 1.872 & 0.662 & 0.841 & 0.638 & 0.729 & 0.774 \\
Ours & \textbf{2.148} & \textbf{1.095} & \textbf{2.322} & \textbf{1.880} & \textbf{1.524} & \textbf{0.719} & \textbf{0.889} & \textbf{0.697} & \textbf{0.791} & \textbf{0.807} \\
\bottomrule[1.2pt]
\end{tabular}
\caption{Results on ShapeNet-ViPC in terms of $L_2$ CD $\times 10^{-3}$ (lower is better) and F-Score@0.001 (higher is better).}
\label{vipc_cd_fs}
\end{table*}

\subsection{Experimental Settings}
We implement the Diffusion-based Coarse Generator (DCG) with batch size as 128 on a single 4090 GPU for 400 to 600 thousand iterations. The dimension of latent code is 512 and the number of reverse steps is 500. The Context-aware Refiner (CRef) is trained for 50 epochs on two 4090 GPUs with batch size as 12. The patch size, the number of generated points, and the number of points in similarity modeling are 64, 2048, 64, respectively. We train a model for all classes on ShapeNet-ViPC and MVP datasets. On PartNet and 3DEPN datasets, models are trained for single classes.
\subsection{Datasets and Metrics}
The comparison is performed on four datasets, including ShapeNet-ViPC \cite{Zhang-vipc}, PartNet \cite{PartNet}, 3D-EPN \cite{3DEPN}, and MVP \cite{VRCNet}.
\textbf{ShapeNet-ViPC} consists of 38,328 objects in 13 categories. Each ground truth is paired with 24 partial point clouds. We utilize the same dataset partitioning method as ShapeNetViPC, employing 31,650 objects from eight categories for all experiments. The dataset is split into 80\% for training and 20\% for testing.
\textbf{3D-EPN} is derived from ShapeNet, providing simulated partial scans with varying levels of incompleteness. We use the provided point cloud  representations in chair, airplane and table categories. For each input, we start with 1,024 points as the partial input and output 2,048 points as the completed shape.
\textbf{PartNet} is a comprehensive, large-scale dataset containing 573,585 part instances across 26,671 3D models spanning 24 object categories. We train our model using Chair, Table, and Lamp categories. For each 1,024 partial input, we output 2,048 points as the completed shape. We use the same dataset partitioning method as cGAN \cite{cGAN}.
\textbf{MVP} consists of 104,000 objects in 16 categories. For each object, there are 26 partial point clouds generated by selecting 26 camera poses and one ground truth point cloud. We use 2048 points for ground truth and follow the dataset partitioning method as PDR \cite{Lyu-Cdiffusion}: 62,400 for training and 41,600 for testing.

We use four standard metrics to evaluate our method. 
\textbf{$L_2$ Chamfer Distance (CD)} is a widely adopted metric, averaging the squared distances between each point in one point cloud and its closest counterpart in the other. It measures the similarity between the completed point cloud and ground truth. 
\textbf{F-Score} is widely used to assess both the accuracy and completeness of a completed point cloud compared to a reference ground truth point cloud. 
\textbf{Earth Mover's Distance (EMD)} assesses how closely a generated point cloud resembles a ground truth point cloud by calculating the minimal cost required to rearrange one point cloud to match the other.
\textbf{Unidirectional Hausdorff Distance (UHD)} calculates the average of the distances from each point in the partial point cloud to the nearest point in the completed one. It measures the completion fidelity with respect to the partial input.
\begin{table}[]
\centering
\begin{tabular}{c|c|ccc}
\toprule[1.2pt]
Methods & Average & Chair & Plane & Table \\
\midrule[0.5pt]
KNN-latent & 1.54 & 1.45 & 0.93 & 2.25 \\
cGAN \shortcite{cGAN} & 1.67 & 1.61 & 0.82 & 2.57 \\
Diverse \shortcite{DSC}& 1.07 & 1.16 & 0.59 & \textbf{1.45} \\
Ours & \textbf{1.04} & \textbf{0.94} & \textbf{0.58} & {1.61} \\
\bottomrule[1.2pt]
\end{tabular}
\caption{Results on 3D-EPN dataset in terms of $L_2$ Chamfer Distance $\times 10^{-3}$ (lower is better)}
\label{3DEPN}
\end{table}
\subsection{Comparison Study}

The comparisons are made between the proposed method and state-of-the-art point cloud completion techniques, with both quantitative and qualitative results being reported.
\begin{table*}[]
\centering
\begin{tabular}{c|cccc|cccc}
\toprule[1.2pt]
\multicolumn{1}{c|}{\multirow{2}{*}{Method}} & \multicolumn{4}{c|}{MMD} & \multicolumn{4}{c}{UHD} \\
\multicolumn{1}{c|}{} & Chair & Lamp & Table & \multicolumn{1}{c|}{Avg.} & Chair & Lamp & Table & Avg. \\
\midrule[0.5pt]
cGAN \shortcite{cGAN} & {1.52} & {1.97} & {1.46} & {1.65} & {6.89} & {5.72} & {5.56} & {6.06} \\
IMLE \shortcite{Huang-PF-Net} & {-} & {-}& {-} & {-}  & {6.17} & {5.58}& {5.16} & {5.64}\\ 
ShapeFormer \shortcite{Yang-FoldingNet} & {-} & {-} & {-} & \textbf{1.32} & {-} & {-} & {-} & {-} \\  
Diverse \shortcite{DSC} & {1.50} & {1.84} & \textbf{1.15} & {1.49} & {3.79} & {3.88} & {3.69} & {3.79} \\  
Ours & \textbf{1.37} & \textbf{1.77} & {1.41} & {1.51} & \textbf{2.55} & \textbf{1.87} & \textbf{2.25} & \textbf{2.22} \\
\bottomrule[1.2pt]
\end{tabular}
\caption{Results on PartNet dataset in terms of MMD $\times 10^{-3}$ (lower is better) and UHD $\times 10^{-2}$ (lower is better).}
\label{PartNet_CD}
\end{table*}
\subsubsection{Results on Shapenet-ViPC.}
On ShapeNet-ViPC, we train our model on all eight categories and evaluate the CD and F-Score@0.001 on each category. We use squared distance when calculate F-Score, following the protocol in ViPC \cite{Zhang-vipc} and CSDN \cite{Zhu-CSDN}.The comparison results are shown in Table \ref{vipc_cd_fs}. Our method achieves the best performance over all methods in all categories. Compared to the second-ranked CSDN, our method reduces the CD by 0.674 (26.4\%) in the lamp category and 0.422 (16.4\%) in average. As for F-Score, our method improves it by 0.031 (5.57\%) in the sofa category and 0.024 (3.45\%) in average. The results of CD and F-score demonstrate that our method achieves enhanced completion performance.

In Figure \ref{figure:visual}, we conduct qualitative comparison on ShapeNet-ViPC in eight categories. The visual results show that our method generates high-fidelity completion results while preserving the high-quality information inherent in partial point clouds.
 
\begin{table}[]
\centering
\begin{tabular}{c|ccc}
\toprule[1.2pt]
{Methods} & {CD} & {EMD} & {F-Score} \\ \midrule[0.5pt]
PCN \shortcite{Yuan-PCN} & {8.65} & {1.95} & {0.342} \\
FoldingNet \shortcite{Yang-FoldingNet} & {10.54} & {3.64} & {0.256} \\
TopNet \shortcite{Tchapmi-TopNet} & {10.19} & {2.44} & {0.299} \\ 
GRNet \shortcite{Xie-GRNet} & {7.61} & {2.36} & {0.353} \\  
VRCNet \shortcite{VRCNet} & {5.82} & {2.31} & {0.495} \\ 
PMPNet++ \shortcite{PMPNet} & {5.85} & {3.42} & {0.475} \\ 
PDR \shortcite{Lyu-Cdiffusion} & {5.66} & \textbf{1.37} & {0.499} \\ 
Ours & \textbf{5.49} & {2.22} & \textbf{0.515} \\
\bottomrule[1.2pt]
\end{tabular}
\caption{Results on MVP in terms of $L_2$ CD $\times 10^{-4}$, EMD $\times 10^{-2}$ and F-Score@0.01.}
\label{MVP_cd}
\end{table}
\subsubsection{Results on 3D-EPN.}
On 3D-EPN, we train a specific model for each category, and evaluate the CD metric on them. In single-class training, our method achieves outstanding performance. The comparison results are shown in Table \ref{3DEPN}. Our method outperforms all competitors in chair and plane categories. Compared to the second-ranked method, we reduces the CD by 0.22 (19.0\%) in the chair category and 0.03 (2.8\%) in average.

\subsubsection{Results on PartNet.}
On PartNet dataset, we train a specific model for each category to evaluate MMD and UHD metrics. We calculate the MMD with CD, following cGAN \cite{cGAN}. In single-class training, our method achieves outstanding performance. As shown in Table \ref{PartNet_CD}, our method achieves the best results in chair and lamp categories, reducing the MMD by 0.13 (8.7\%) and 0.07 (3.8\%) respectively. As for UHD, we achieve the best results in all categories, reducing the UHD by 1.24 (32.7\%), 2.01 (51.8\%), and 1.44 (39.0\%) respectively. In average, the reduction is 1.57 (41.4\%). Results demonstrates that our method achieves high-fidelity completion.

\subsubsection{Results on MVP.}
We train our model on all 16 categories and report the CD and F-Score@0.01 metrics. Although the MMD result is not the best, our method reduces the CD by 0.17 (3.0\%) and improves the F-Score by 0.016 (3.2\%). 
\subsection{Method Analysis}
\begin{figure}[t]
\centering
\includegraphics[width=0.999\columnwidth]{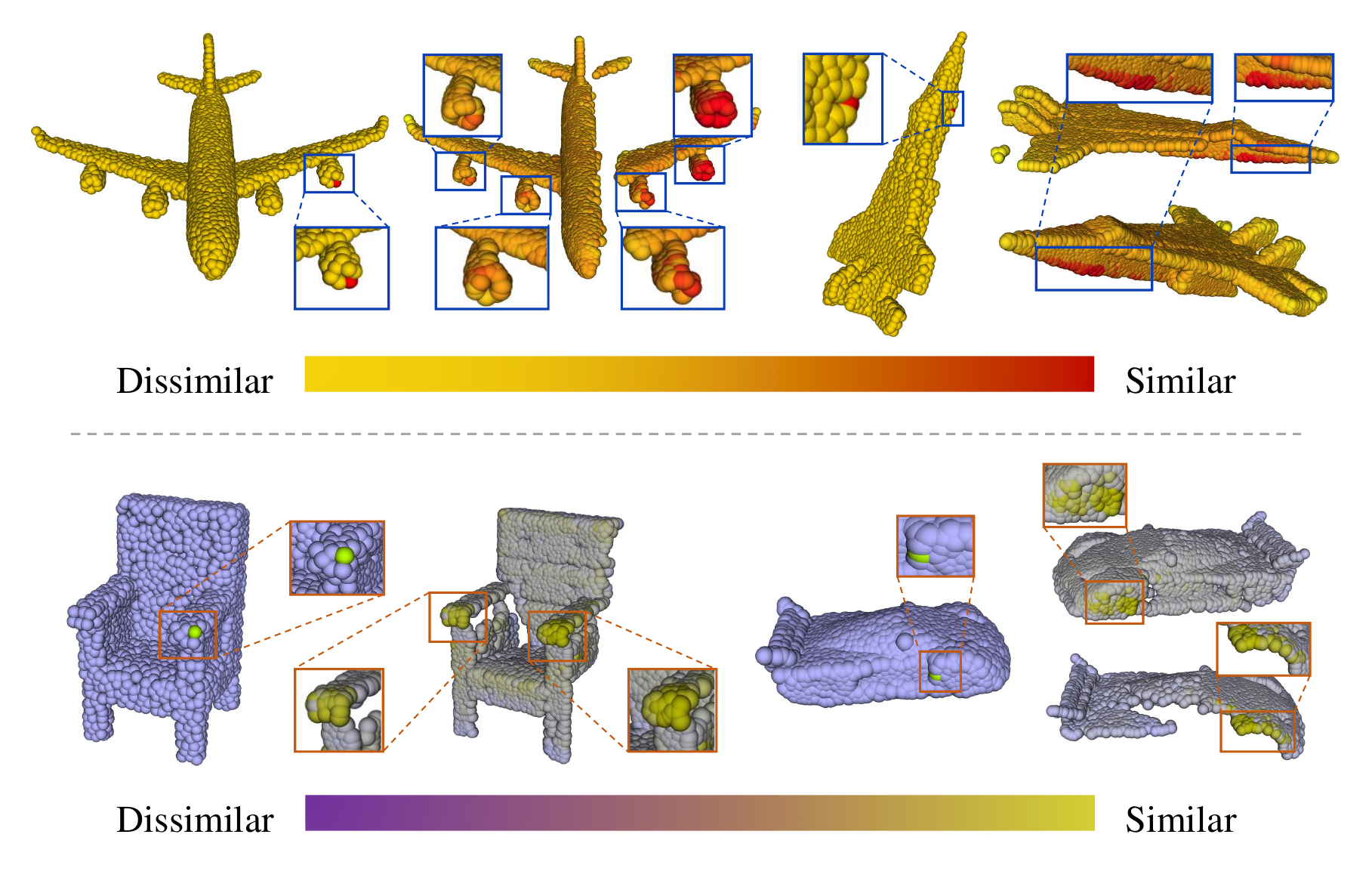} 
\caption{For each pair of images, the left image highlights a specific point within the complete point cloud. The accompanying heatmap on the right displays the similarity of each point in the partial point cloud to that reference point. A higher degree of similarity is indicated by more intense colors: red for airplanes and yellow for chairs and cars.}
\label{figure:similarity}
\end{figure}
We conduct ablation study and similarity study to demonstrate the effectiveness of the several proposed operations.
\subsubsection{Ablation Study}
\begin{table}[]
\centering
\setlength{\tabcolsep}{1.2mm}{\begin{tabular}{c|c|ccc}
\toprule[1.2pt]
{Methods} & {Avg.} & {Airpalne} & {Car} & {Chair} \\ \midrule[0.5pt]
W/o mixed sampling & {2.96} & {1.44} & {3.01} & {2.69} \\
W/o surface freezing & {2.74} & {1.31} & {2.83} & {2.61} \\ 
W/o rigid transformation & {2.43} & {1.29} & {2.95} & {2.57} \\
Euclidean similarity & {2.24} & {1.20} & {2.84} & {2.53}\\  
Feature similarity & {2.22} & {1.18} & {2.82} & {2.51} \\ 
Ours & \textbf{2.21} & \textbf{1.12} & \textbf{2.82} & \textbf{2.50} \\
\bottomrule[1.2pt]
\end{tabular}}
\caption{Different experiments for ablation study in terms of $L_2$ Chamfer Distance $\times 10^{-3}$ (lower is better).}
\label{ablation}
\end{table}
We conduct five ablation experiments on a slice of shapeNet-ViPC dataset, as shown in Table \ref{ablation}. 

CRef w/o mixed sampling. We remove the mixed sampling module in the Context-aware Refiner (CRef). This results in inaccuracies in short-range contextual information, which may subsequently impact the experimental results. 

CRef w/o surface freezing. We remove the surfacing freezing module in CRef. This causes the displacement of points located precisely on the lower surface, which may consequently impact the experimental results.

CRef w/o rigid transformation. We remove the rigid transformation in CRef. The features extracted by the network do not encompass those invariant to rigid transformations, thereby affecting the completion results.

Euclidean similarity. We use euclidean distance only in the similarity modeling. During the refinement stage, considering only the features of the points surrounding them leads to a decline in the quality of the completion.

Feature similarity. We use euclidean distance only in the similarity modeling. During the refinement stage, referring to only the points near them in feature space leads to a little decline in the quality of the completion.

\subsubsection{Similarity Study}
To test the effectiveness of our non-local similarity modeling operation, we visualize the non-local similarity (learned by our model) between complete point cloud and the partial one. This point-wise similarity captures the extent to which the geometric manifold structures around the points are alike. Figure \ref{figure:similarity} shows that our method effectively learns a robust matching of non-local similarity.
\section{Conclusions}
We propose a high-fidelity point cloud completion method with a two-stage structure to digs into both short-range and long-range contextual information. We design a mixed sampling module and surface freezing mechanism to incorporate short-range contextual details and a rigid transformation-invariant feature extractor to extract long-range contextual information. Extensive comparisons and ablation studies are conducted to demonstrate the effectiveness of our method.
\section{Acknowledgments}
This work was supported in part by the National Key R\&D Program of China (2021YFF0900500), the National Natural Science Foundation of China (NSFC) under grants 62441202, U22B2035, 20240222, and the Fundamental Research Funds for the Central Universities under grants HIT.DZJJ.2024025.
\bibliography{aaai25}
\clearpage
\end{document}